\pdfoutput=1

\documentclass[11pt]{article}

\usepackage[final]{acl}

\usepackage{times}
\usepackage{latexsym}

\usepackage[T1]{fontenc}

\usepackage[utf8]{inputenc}

\usepackage{microtype}

\usepackage{inconsolata}

\usepackage{graphicx}
\usepackage{float}
\usepackage{amsfonts}
\usepackage{amsmath}
\usepackage{hyperref}
\usepackage{url}
\usepackage{multirow}
\usepackage{caption}
\usepackage{subcaption}
\usepackage{booktabs}
\usepackage{nicefrac}
\usepackage{microtype}
\usepackage{xcolor}
\usepackage{multirow}
\usepackage{epsfig}
\usepackage{amsmath}
\usepackage{diagbox}
\usepackage{float}
\usepackage{pifont}
\usepackage{color}
\usepackage{xspace}
\usepackage{caption}
\usepackage{wrapfig}
\usepackage{varwidth}

\usepackage{color}
\usepackage{times}
\usepackage{latexsym}
\usepackage{graphicx}
\usepackage[T1]{fontenc}
\usepackage[utf8]{inputenc}
\usepackage{microtype}
\usepackage{inconsolata}
\usepackage{amsmath}

\usepackage{multirow} 
\usepackage{array}
\usepackage{xcolor, soul}
\usepackage{arydshln}
\usepackage{makecell}

\usepackage{tcolorbox}

\usepackage{algorithm}
\usepackage{algpseudocode}

\usepackage{tikz}

\newcolumntype{M}{>{\begin{varwidth}{4cm}}l<{\end{varwidth}}}

\newcommand*{\rowstyle}[1]{
  \gdef\@rowstyle{#1}%
  \@rowstyle\ignorespaces%
}
\newcolumntype{=}{
  >{\gdef\@rowstyle{}}%
}
\newcolumntype{+}{
  >{\@rowstyle}%
}
\definecolor{revcolor}{rgb}{0,0,0}

\newcounter{magicrownumbers}

\title{Learning to Reason Over Time: Timeline Self-Reflection for Improved Temporal Reasoning in Language Models}

\author{Adrián Bazaga\thanks{Work done during an internship at Amazon. Now at Microsoft.}
\\
  University of Cambridge\\
  \texttt{ar989@cam.ac.uk} \\\And
  Rexhina Blloshmi \\
  Amazon AGI \\ 
  \texttt{blloshmi@amazon.com} \\\AND
  Bill Byrne \\
  Amazon AGI \\
  \texttt{willbyrn@amazon.com} \\\And
  Adrià de Gispert \\
  Amazon AGI \\
  \texttt{agispert@amazon.com} \\}

\begin{document}

\maketitle

\begin{abstract} 
Large Language Models (LLMs) have emerged as powerful tools for generating coherent text, understanding context, and performing reasoning tasks. However, they struggle with temporal reasoning, which requires processing time-related information such as event sequencing, durations, and inter-temporal relationships. These capabilities are critical for applications including question answering, scheduling, and historical analysis. In this paper, we introduce TISER, a novel framework that enhances the temporal reasoning abilities of LLMs through a multi-stage process that combines timeline construction with iterative self-reflection. Our approach leverages test-time scaling to extend the length of reasoning traces, enabling models to capture complex temporal dependencies more effectively. This strategy not only boosts reasoning accuracy but also improves the traceability of the inference process. Experimental results demonstrate state-of-the-art performance across multiple benchmarks, including out-of-distribution test sets, and reveal that TISER enables smaller open-source models to surpass larger closed-weight models on challenging temporal reasoning tasks.\footnote{Code and data are available at \href{https://github.com/amazon-science/TISER}{https://github.com/amazon-science/TISER}.} 
\end{abstract}

\section{Introduction}

Temporal reasoning, defined as the ability to interpret and manipulate time-related information, is a fundamental aspect of human cognition that underpins language understanding, planning, forecasting, and the interpretation of sequences, durations, and temporal relationships~\cite{vashishtha-etal-2020-temporal, huang-chang-2023-towards, minetal2023nlpadvances}. In natural language processing (NLP), effective temporal reasoning is critical for applications such as question answering (QA)~\cite{zhao2024surveylargelanguagemodels}, event prediction~\cite{Dhingra2021TimeAwareLM}, narrative understanding~\cite{minetal2023nlpadvances}, and retrieval-augmented generation (RAG)~\cite{rag2020}. However, despite the remarkable advances of large language models (LLMs) in a wide range of tasks, achieving accurate and robust reasoning over temporal contexts remains a significant challenge~\cite{zhou-etal-2019-going, wang-zhao-2024-tram, su2024timobettertemporalreasoning, qiu-etal-2024-large}.

Recent benchmarks, such as TRAM~\cite{wang-zhao-2024-tram} and TimeBench~\cite{chu2024timebenchcomprehensiveevaluationtemporal}, have underscored the difficulties LLMs face in handling complex temporal queries. In many cases, even state-of-the-art models produce inconsistent inferences or fail to accurately track the sequential order of events. Traditional approaches to improving temporal reasoning have relied on prompt engineering~\cite{wei2023specific, zhang-etal-2024-narrative}, supervised fine-tuning~\cite{li2023unlocking, qian-etal-2024-timer4}, specialized pre-training~\cite{tan-etal-2023-towards}, or mathematical reasoning modules~\cite{su2024timobettertemporalreasoning}.

Inspired by recent advancements in test-time scaling of LLMs for reasoning tasks~\cite{muennighoff2025s1simpletesttimescaling,deepseekai2025deepseekr1incentivizingreasoningcapability}, we propose a novel paradigm for adapting LLMs to temporal reasoning through test-time scaling. Our framework, TISER, incorporates a multi-stage inference pipeline that combines explicit reasoning, timeline construction, and iterative self-reflection. The key idea behind our approach is to empower LLMs to adapt by scaling their internal reasoning process during inference. TISER enables models to systematically organize temporal information, verify their inferences, and refine their outputs. This approach not only enhances temporal reasoning accuracy but also improves traceability by making the intermediate reasoning steps explicit. Our contributions are as follows:

\begin{itemize} 
\item We introduce TISER, a novel framework that adapts LLMs for test-time temporal reasoning via a multi-stage process encompassing reasoning, explicit timeline construction, and self-reflection.

\item We construct a synthetic dataset that augments existing temporal reasoning benchmarks with detailed intermediate reasoning traces, enabling effective test-time adaptation.

\item We demonstrate that fine-tuning and test-time scaling using TISER achieves state-of-the-art temporal reasoning performance on multiple benchmarks, including TGQA~\cite{xiong-etal-2024-large}, TempReason~\cite{tan-etal-2023-towards}, and TimeQA~\cite{chen2021dataset}. 

\item We evaluate the fine-tuned LLMs on out-of-distribution benchmarks such as MultiHopRAG~\cite{tang2024multihoprag} and Test-of-Time (ToT)~\cite{fatemi2024testtimebenchmarkevaluating}, demonstrating that our approach preserves performance on standard queries while boosting accuracy on time-based ones.
\end{itemize}

\section{TISER}

\begin{figure*}[th!]
\centering
\includegraphics[width=\textwidth]{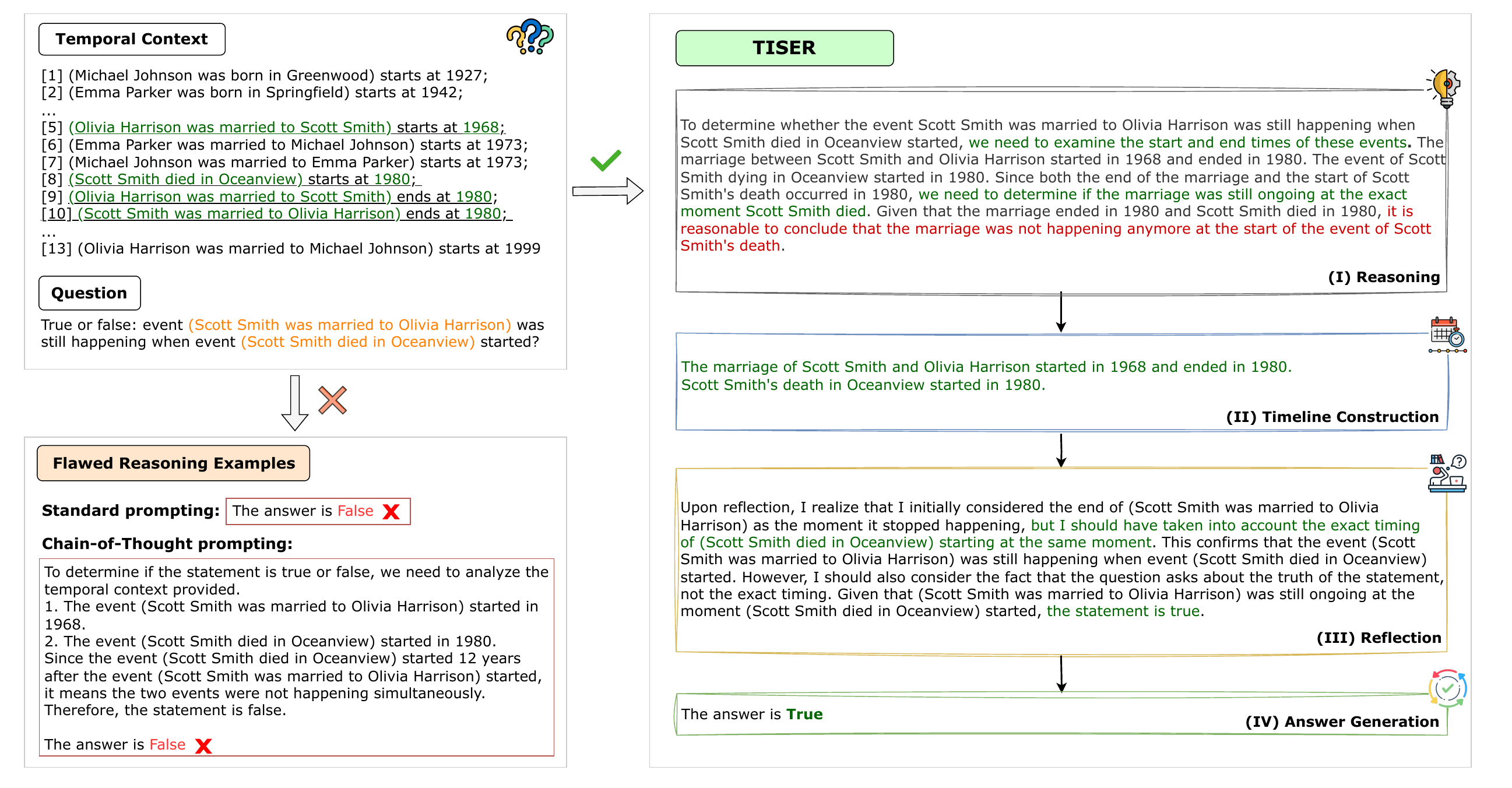}
\caption{High-level overview of TISER (right) compared to other prompting strategies such as standard prompting or CoT without test-time scaling (bottom left) for a given question and temporal context (upper left). In contrast to standard prompting and CoT, our method leverages test-time compute scaling for reasoning, timeline construction, and reflection, leading to more accurate answers. The model used for inference in this example is Qwen2.5-7B.}
\label{fig:tiser_method}
\end{figure*}

\subsection{Temporal Self-Reflective Prompting}

In order to enhance temporal reasoning in large language models (LLMs), we introduce Temporal Self-Reflective Prompting (TISER). This strategy advances traditional Chain-of-Thought (CoT) prompting by explicitly decomposing the temporal reasoning process into multiple stages, each designed to progressively refine the model’s internal representation and final output. The key innovation of TISER lies in its integration of explicit timeline construction and self-reflection for test-time scaling, which collectively mitigate typical errors in temporal inference and improve both the traceability and robustness of the reasoning process.

As detailed in Algorithm~\ref{alg:tiser}, TISER processes a given question $q$ and its associated temporal context $c$ through four interdependent stages, each building upon the previous one:

\begin{algorithm}[t!]
\caption{Temporal Self-Reflective Prompting} 
\label{alg:tiser} 
\textbf{Input:} Question $q$, Temporal context $c$\\
\textbf{Output:} Final answer $a$ 
    
\begin{algorithmic}[1] 
  \State \textbf{// Stage I: Reasoning} 
  \State Generate an initial reasoning trace $r$ for $q$ based on context $c$. 
  \State 
  \Repeat
    \State \textbf{// Stage II: Timeline Construction} 
    \State Extract relevant temporal events from $r$ and $c$. 
    \State Organize the extracted events into an ordered timeline $t$. 
    \State 
    \State \textbf{// Stage III: Reflection} 
    \State Analyze $r$ in conjunction with timeline $t$ to detect inconsistencies or errors. 
    \State Revise the reasoning trace to produce an improved version $r'$. 
    \State 
    \State \textbf{Set} $r \gets r'$
  \Until{the revised reasoning $r'$ and timeline $t$ are consistent and satisfactory}
  \State 
  \State \textbf{// Stage IV: Answer Generation} 
  \State Generate the final answer $a$ using the refined reasoning $r'$ and timeline $t$. 
  \State \Return $a$ 
\end{algorithmic} 
\end{algorithm}

\textbf{Stage I (Reasoning):} The model initiates the process by generating a preliminary chain-of-thought reasoning trace, denoted as $r$, based on the question $q$ and the temporal context $c$. This initial reasoning encapsulates the model's raw inference process and prepares the groundwork for further refinement.

\textbf{Stage II (Timeline Construction):} In this stage, the algorithm identifies and extracts salient temporal events from both the reasoning trace $r$ and the context $c$. These events are then organized into a coherent, ordered timeline $t$, which serves as an explicit representation of the temporal structure inherent in the problem. By grounding the reasoning in a structured timeline, the model can more clearly delineate the sequence and interdependencies of events.

\textbf{Stage III (Reflection):} The model then engages in a self-reflective process, comparing the initial reasoning trace $r$ with the constructed timeline $t$. This stage is critical for detecting and rectifying inconsistencies, ambiguities, or omissions in the initial reasoning. The reflection yields a refined reasoning trace $r'$, representing an improved and self-verified version of the original inference.

\textbf{Stage IV (Answer Generation):} Finally, leveraging the refined reasoning $r'$ and the explicit timeline $t$, the model generates the final, concise answer $a$. This final output is expected to be both accurate and logically coherent, as it has benefited from iterative error correction and temporal grounding.

By explicitly segmenting the reasoning process into these four interdependent stages, TISER allows for iterative improvement and facilitates a more robust understanding of complex temporal relationships, ultimately leading to more precise and reliable answers.
In Figure~\ref{fig:tiser_method} we illustrate an end-to-end example with TISER reasoning.

\subsection{Dataset Construction}
\label{data-construction}

To adapt our models for test-time scaling on temporal reasoning tasks, we construct a high-quality dataset that augments existing temporal reasoning datasets with intermediate reasoning traces, which we use for fine-tuning. Each sample in the dataset comprises a question $q$, its corresponding gold-standard answer $a$, and a temporal context $c$. 
We generate intermediate reasoning steps using any off-the-shelf LLM for synthetic data generation, such as DeepSeek or GPT-4o, guided by the TISER prompt framework to ensure consistent response formatting\footnote{The full TISER prompt used for dataset generation is shown in Appendix~\ref{app:tiser_prompt_example}.}. In particular, the generated traces include a reasoning sequence $r$, an ordered timeline of events $t$, and a reflective verification $f$. Since each question is paired with a verified gold answer, the generator LLM is encouraged to reason through problems using a temporally consistent process to arrive at the final correct `gold' answer. 

To ensure dataset quality, we apply a filtering procedure that retains only those samples for which the final answer $a'$ produced from the intermediate reasoning trace matches the gold answer $a$. Samples that do not pass this consistency check are discarded. The overall dataset construction process is summarized in Algorithm~\ref{alg:dataset}. Detailed quality metrics for the generated dataset are provided in Appendix~\ref{app:dataset_quality_stats}.

\begin{algorithm}[t!]
\caption{Dataset Construction with Intermediate Reasoning Traces}
\label{alg:dataset}

\textbf{Input:} Collection of samples \(\mathcal{D}\) containing \((q, a, c)\) from an existing temporal reasoning dataset.\\
\textbf{Output:} Augmented dataset \(\mathcal{D}^*\) with intermediate reasoning traces.
\begin{algorithmic}[1]
\For{each sample \((q, a, c) \in \mathcal{D}\)}
    \State Generate an intermediate reasoning trace containing: reasoning $r$, timeline $t$, and reflection $f$.
    \State Given $r$, $t$ and $f$, generate an answer $a'$ to the question $q$ with the context $c$.
    \If{\(a' = a\)}
         \State Add sample \((q, a, c, r, t, f)\) to \(\mathcal{D}^*\).
    \Else
         \State Discard sample.
    \EndIf
\EndFor\\
\Return \(\mathcal{D}^*\)
\end{algorithmic}
\end{algorithm}

\subsection{Fine-tuning for Temporal Reasoning}

We employ a standard supervised fine-tuning (SFT) procedure using Low-Rank Adaptation (LoRA)~\cite{hu2022lora} to finetune the base models. We train our models using the dataset constructed according to the TISER framework described in Section~\ref{data-construction}. Specifically, the generated outputs adhere to a defined template: a step-by-step chain-of-thought is enclosed within \texttt{<reasoning>} tags, an ordered summary of relevant temporal events is provided within \texttt{<timeline>} tags, the model's self-reflection and verification are captured within \texttt{<reflection>} tags, and the final concise answer is delimited by \texttt{<answer>} tags. Further details regarding the fine-tuning process are provided in Appendix~\ref{app:detailed_finetuning}.

\begin{table*}[!t]
\centering
\resizebox{\textwidth}{!}{%
\begin{tabular}{llcccccccccccc}
\toprule
\textbf{Model} & \textbf{Inference} & \multicolumn{2}{c}{\textbf{TGQA}} & \multicolumn{2}{c}{\textbf{TempReason (L2)}} & \multicolumn{2}{c}{\textbf{TempReason (L3)}} & \multicolumn{2}{c}{\textbf{TimeQA (easy)}} & \multicolumn{2}{c}{\textbf{TimeQA (hard)}} & \multicolumn{2}{c}{\textbf{Macro Avg.}} \\ 
\cmidrule(lr){3-4} \cmidrule(lr){5-6} \cmidrule(lr){7-8} \cmidrule(lr){9-10} \cmidrule(lr){11-12} \cmidrule(lr){13-14}
  &  & \textbf{EM} & \textbf{F1} & \textbf{EM} & \textbf{F1} & \textbf{EM} & \textbf{F1} & \textbf{EM} & \textbf{F1} & \textbf{EM} & \textbf{F1} & \textbf{EM} & \textbf{F1}  \\ 
\midrule
\multicolumn{13}{l}{\textbf{Closed LLMs}} \\
\midrule
GPT-4 & Standard & 72.5 & 82.5 & 78.6 & 86.2 & 81.9 & 88.3 & 83.6 & 93.7 & 76.0 & 85.3 & 78.5 & 87.2 \\ 
GPT-4 & TISER & 82.8 & 93.4 & 79.8 & 87.2 & 84.7 & 91.3 & 84.4 & 90.5 & 77.2 & 86.4 & 81.8 & 89.8 \\ 
\midrule
\multicolumn{13}{l}{\textbf{Literature}} \\
\midrule
T5-large$^{\dag}$ & Standard & 54.8 & 71.3 & 32.7 & 50.9 & 28.8 & 46.8 & 63.1 & 71.6 & 59.5 & 68.1 & 47.8 & 61.7 \\ 
Temp-T5$^{\dag}$ & Standard & 64.0 & 77.8 & 31.8 & 49.6 & 26.1 & 43.0 & -- & -- & -- & -- & -- & -- \\ 
REMEMO-large$^{\dag}$ & Standard & 46.1 & 66.0 & 37.4 & 54.9 & 33.4 & 49.3 & 63.7 & 72.3 & 60.5 & 69.3 & 48.2 & 62.4 \\ 
TG-LLM$^{\dag}$ & CoT & 79.7 & 85.0 & 42.4 & 52.2 & 35.6 & 46.9 & 66.4 & 69.1 & 63.1 & 66.4 & 57.4 & 63.9\\ 
\midrule
\multicolumn{13}{l}{\textbf{Open LLMs - Off-the-Shelf}} \\
\midrule
Mistral-7B & Standard & 15.1 & 18.6 & 22.0 & 24.1 & 0.1 & 2.3 & 51.8 & 55.2 & 37.0 & 39.0 & 25.2 & 27.8 \\ 
Qwen2.5-7B & Standard & 46.1 & 48.9 & 51.0 & 53.6 & 40.1 & 42.7 & 70.9 & 73.5 & 53.2 & 55.8 & 52.3 & 55.0 \\ 
Mistral-7B & TISER & 22.4 & 24.7 & 20.8 & 23.6 & 36.8 & 38.9 & 43.3 & 46.1 & 22.1 & 25.3 & 29.1 & 31.7 \\ 
Qwen2.5-7B & TISER & 48.1 & 50.6 & 55.2 & 58.2 & 51.6 & 54.2 & 66.1 & 68.6 & 55.2 & 57.7 & 55.3 & 57.9 \\ 

\midrule
\multicolumn{13}{l}{\textbf{Open LLMs - Fine-tuned} \textit{(training data using standard prompting)}} \\
\midrule
Mistral-7B & Standard & 18.3 & 21.2 & 25.4 & 27.5 & 9.3 & 12.8 & 55.6 & 59.1 & 41.2 & 44.3 & 31.7 & 34.5 \\ 
Qwen2.5-7B & Standard & 48.6 & 51.7 & 54.0 & 56.6 & 44.2 & 46.7 & 74.1 & 76.9 & 56.1 & 58.7 & 55.4 & 57.3 \\ 

\midrule
\multicolumn{13}{l}{\textbf{Open LLMs - Fine-tuned} \textit{(TISER training data generated using DeepSeek V2.5)}} \\
\midrule
Mistral-7B & Standard & 48.7 & 52.4 & 49.3 & 51.0 & 60.8 & 63.2 & 61.4 & 62.5 & 59.2 & 60.7 & 55.7 & 57.9 \\  
Qwen2.5-7B & Standard & 53.9 & 56.6 & 49.2 & 51.9 & 39.6 & 41.9 & 68.4 & 70.7 & 51.1 & 53.4 & 52.4 & 54.9 \\ 
Mistral-7B & TISER & 76.3 & 84.0 & 78.5 & 80.1 & 83.0 & 86.6 & 94.2 & 95.2 & 91.4 & 92.5 & 85.6 & 89.9 \\ 
Qwen2.5-7B & TISER & 79.4 & 86.8 & 84.7 & 86.6 & 89.9 & 91.7 & 87.6 & 89.8 & 81.2 & 89.7 & 84.6 & 88.9 \\ 

\midrule
\multicolumn{13}{l}{\textbf{Open LLMs - Fine-tuned} \textit{(TISER training data generated using GPT-4)}} \\
\midrule
Mistral-7B & Standard & 52.9 & 57.0 & 55.9 & 57.2 & 64.3 & 66.7 & 64.4 & 65.5 & 62.9 & 64.0 & 60.1 & 62.1 \\ 
Qwen2.5-7B & Standard & 58.5 & 60.8 & 53.0 & 55.6 & 43.0 & 45.2 & 72.6 & 74.8 & 54.5 & 56.8 & 54.3 & 57.0 \\ 
Mistral-7B & TISER & 80.5 & 87.4 & 82.5 & 84.3 & 87.1 & 88.5 & 97.5 & 98.5 & 95.9 & 96.4 & 88.7 & 91.0 \\ 
Qwen2.5-7B & TISER & \textbf{84.5} & \textbf{94.2} & \textbf{85.5} & \textbf{87.5} & \textbf{91.5} & \textbf{94.9} & \textbf{97.9} & \textbf{98.3} & \textbf{96.1} & \textbf{97.2} & \textbf{91.1} & \textbf{94.4} \\ 

\bottomrule
\end{tabular}%
}
\caption{Exact Match (EM) and token-level F1 results on in-domain test sets. Fine-tuned models are trained on the joint training dataset, while evaluation is done on separate test splits. The Inference column gives the prompting strategy used in evaluation. Results with $^{\dag}$ are reported in the original papers}
\label{tab:main_results}
\end{table*}

\section{Experiments}

We evaluate TISER on a range of closed and open LLMs, with and without fine tuning, on multiple temporal reasoning benchmarks along with a more general RAG downstream task. 

\subsection{Experimental Setup}

\paragraph{Datasets.} The training dataset is derived from a combination of existing temporal reasoning benchmarks, including TGQA\footnote{\url{https://huggingface.co/datasets/sxiong/TGQA/viewer/TGQA_TGR}}~\cite{xiong-etal-2024-large}, TempReason\footnote{\url{https://huggingface.co/datasets/sxiong/TGQA/viewer/TempReason_TGR/}}~\cite{tan-etal-2023-towards}, and TimeQA\footnote{\url{https://huggingface.co/datasets/sxiong/TGQA/viewer/TimeQA_TGR/}}~\cite{chen2021dataset}. We apply our dataset construction method described in Section~\ref{data-construction} to create the final training examples. Fine-tuning was conducted on the individual datasets, as well as a joint dataset combining all of them. We evaluate on the test splits of TGQA, TempReason and TimeQA, which we refer to as in-domain evaluation.
These datasets cover a range of temporal reasoning challenges, such as temporal event ordering and duration calculations. We also evaluate on the ToT benchmark\footnote{\url{https://huggingface.co/datasets/baharef/ToT}}~\cite{fatemi2024testtimebenchmarkevaluating} and MultiHopRAG\footnote{\url{https://huggingface.co/datasets/yixuantt/MultiHopRAG}}~\cite{tang2024multihoprag} to assess the benefits of TISER in the out-of-distribution scenario and QA downstream tasks. Details and examples are in Appendix~\ref{app:dataset_description_examples}.

\paragraph{Metrics.} We evaluate models using Exact Match (EM), which measures the percentage of predictions that exactly match the ground truth, and token-level F1 scores. These metrics are particularly suited for temporal reasoning tasks, where precision is critical for accurate event alignment and inference.

\paragraph{Models.} Our experiments compare the performance of closed LLMs (e.g., GPT-4o) and open LLMs (e.g., Mistral-7B, Qwen2.5-7B) under different settings: 1) using standard prompts, i.e., baseline performance using the original prompts from each test set, without reasoning~(see Appendix~\ref{app:standard_prompt_example} for an example), 2) using TISER prompt, i.e., applying our proposed reasoning, timeline construction and self-reflection strategy during inference only or also fine-tuning. We include both off-the-shelf and fine-tuned configurations of open-source models to assess the relative contributions of TISER during fine-tuning and inference. 

\begin{table*}[t]
\centering
\resizebox{1\linewidth}{!}{
\begin{tabular}{lcccccc}
\toprule
\textbf{Training subset} & \textbf{TGQA} & \textbf{TempReason (L2)} & \textbf{TempReason (L3)} & \textbf{TimeQA (easy)} & \textbf{TimeQA (hard)} & \textbf{Macro Avg.} \\ 
\midrule
TGQA              & 80.5               & 61.5                     & 77.9                     & 76.1                  & 63.5                  & 72.5                  \\ 
TempReason (L2)    & 55.1               & 81.5                     & 80.5                     & 82.5                  & 75.5                  & 74.6                  \\ 
TempReason (L3)    & 50.1               & 81.0                     & 81.7                     & 80.1                  & 70.9                  & 73.2                  \\ 
TimeQA (easy)      & 46.3               & 72.4                     & 73.4                     & 92.8                  & 78.6                  & 72.5                  \\ 
TimeQA (hard)      & 47.8               & 81.3                     & 75.5                     & 93.0                  & 91.1                  & 77.6                  \\ 
\cmidrule[0.5pt]{1-7}
all (TISER dataset)      & \textbf{84.5}      & \textbf{85.5}            & \textbf{91.5}            & \textbf{97.9}         & \textbf{96.1}         & \textbf{91.1}         \\ 
\bottomrule
\end{tabular}}
\caption{Exact Match (\%) results after fine-tuning Qwen2.5-7B independently on different training subsets (first column) using TISER prompt. Bold text indicates best performance for that particular test set (column).}
\label{tab:training_subsets_vs_joint}
\end{table*}

\paragraph{Implementation details.}
We use Qwen2.5-7B and Mistral-7B as our main open models, and GPT-4o\footnote{GPT-4o, version \textit{gpt-4o-2024-08-06}} via API for closed-model evaluations. Open models are fine-tuned via SFT using the Hugging Face Transformers library on 8x NVIDIA A100 GPUs. As discussed in Section~\ref{data-construction}, the TISER dataset can be constructed using any off-the-shelf LLM. For our experiments, we evaluate GPT-4o and DeepSeek V2.5 as generators for the self-reflection examples of temporal reasoning.

\subsection{Main Results}
\paragraph{Effectiveness of TISER Prompting across models.} TISER prompting enables off-the-shelf models to tackle complex temporal reasoning tasks more effectively in many cases, with substantial improvements in several benchmarks. 
As shown in Table~\ref{tab:main_results}, Qwen2.5-7B shows a notable increase in accuracy from 51.0\% to 55.2\% on TempReason (L2) and from 40.1\% to 51.6\% on TempReason (L3) when using TISER prompting compared to standard prompting. This improvement highlights the value of TISER prompting in enhancing reasoning and temporal coherence in models that otherwise struggle with these tasks. However, TISER prompting itself does not consistently improve performance in all datasets. For example, Mistral-7B achieves a higher accuracy in TempReason (L2) with standard prompting (22.0\%) compared to TISER prompting (20.8\%). This discrepancy may be due to the fact that these models are not specifically trained to follow the self-reflection framework proposed in TISER, so they cannot use TISER prompting to improve reasoning and temporal alignment in their responses. These results suggest that even without fine-tuning, TISER prompting can enhance performance in certain contexts. Nevertheless, the benefits of TISER prompting are more consistent and significantly pronounced when models are fine-tuned with the TISER strategy.

\paragraph{Performance of Open LLMs: Baselines vs TISER strategy.} The open LLMs examined in this study show varied baseline performance but all benefit substantially from the application of TISER prompting, particularly when TISER is used during both training (fine-tuning) and inference. For example, Mistral-7B's TGQA score increases from 15.1\% to 80.5\%, and its macro average improves from 25.2\% to 88.7\%, demonstrating how TISER enables smaller models to rival larger ones. Similarly, Qwen2.5-7B achieves a significant increase in accuracy on temporal reasoning-heavy datasets, reaching 85.5\% on TempReason (L2) and 91.5\% on TempReason (L3), with its macro-average going up from 57.9\% to 91.1\%, outperforming competitive models across multiple datasets. In addition, the current state-of-the-art, TG-LLM \cite{xiong-etal-2024-large}, while performing well on datasets such as TGQA (79.7\%), struggles with other temporal reasoning tasks, with accuracy dropping to 42.4\% on TempReason (L2) and 35.6\% on TempReason (L3). The macro average of TG-LLM of 57.4\% underscores its limitations, especially compared to models that leverage TISER prompting. These findings highlight that the integrated training and inference approach of TISER not only increases performance in temporal and reasoning-intensive tasks but also enables the models to handle a wider range of complexities more effectively than current models.

\paragraph{Comparison of Closed vs. Open LLMs.} In their baseline configurations, closed models like GPT-4o generally outperform open models like Mistral-7B and Qwen2.5-7B, especially when only standard prompting is applied during inference. However, when TISER is used during both training and inference to fine-tune open models, the performance gap narrows significantly. In some cases, such as in TempReason (L3), Qwen2.5-7B surpasses GPT-4o, reaching 91.5\% compared to GPT-4o’s 84.7\%. This demonstrates the potential for open models to outperform closed models when they are fine-tuned with our proposed strategy.

\paragraph{Performance Across Training Subsets vs Joint Dataset.} Table~\ref{tab:training_subsets_vs_joint} reports the results after independently fine-tuning the Qwen2.5-7B model on different training subsets and evaluating its performance on multiple test sets. Performance is calculated in terms of EM across all test sets, with the best score for each test set highlighted in bold. The training subsets represent individual or combined datasets used for fine-tuning, while the evaluation spans several reasoning datasets, including TGQA, TempReason (L2 and L3), and TimeQA (easy and hard). The macro average column provides an overall performance summary across all datasets. Fine-tuning the model on the combined dataset that includes all subsets produces the best macro average performance, achieving a score of 91.1\%, highlighting the effectiveness of training on a diverse set of reasoning tasks. The joint fine-tuning approach enables the model to achieve best performance for all individual tasks, including 91.5\% in TempReason (L3), 97.9\% in TimeQA (easy) and 96.1\% in TimeQA (hard). These results consistently surpass those achieved through fine-tuning in individual subsets, where performance tended to vary and was often lower across test sets. For example, fine-tuning on individual subsets such as TGQA, TempReason (L2), and TempReason (L3) resulted in lower performance on several test sets, particularly in TGQA and TempReason (L2). Although subset-specific fine-tuning can lead to specialization on certain tasks, such as TempReason (L2) scoring 81.5\% on TempReason (L2), it struggles to generalize effectively across diverse reasoning datasets. Similarly, the TempReason (L3) subset showed moderate performance but fell short compared to the combined dataset, scoring 81.7\% in TempReason (L3) and a macro average of 73.2\%.

\subsection{Out-of-Distribution Results}

Table~\ref{tab:multihoprag_results} showcases a detailed analysis of model performance on the MultiHopRAG benchmark with standard prompting~\footnote{We use the same grounding as in the paper and omit null queries as deflection is out-of-scope for this work.}, comparing multiple models across various question types (Inference, Comparison, and Temporal) using Exact Match accuracy as the metric. 

\begin{table}[t]
\centering
\resizebox{\columnwidth}{!}{%
\begin{tabular}{lccc}
\toprule
\textbf{Model} & \multicolumn{3}{c}{\textbf{Question Type}} \\ 
               & \textbf{Inference} & \textbf{Comparison} & \textbf{Temporal} \\ 
\midrule
\multicolumn{4}{l}{\textbf{Closed LLMs}} \\ 
\midrule
GPT-4         & \textbf{89.7}      & \textbf{77.5}      & \textbf{61.9}       \\ 
\midrule
\multicolumn{4}{l}{\textbf{Open LLMs - Off-the-Shelf}} \\ 
\midrule
Mistral-7B    & 8.7               & 15.9               & 12.0                \\ 
Qwen2.5-7B    & 30.5              & 25.0               & 14.8                \\ 
\midrule
\multicolumn{4}{l}{\textbf{Open LLMs - Fine-tuned} \textit{(TISER data generated using GPT-4)}} \\ 
\midrule
Mistral-7B    & 29.2              & 21.6               & 27.3                \\ 
Qwen2.5-7B    & 31.8              & 30.0               & 33.5                \\ 
\bottomrule
\end{tabular}%
}
\caption{Exact Match (\%) results on a broad-domain RAG task (MultiHopRAG). Performance is reported by question type (inference, comparison, and temporal) using a single standard prompt for all questions vs. a prompt per question type as in MultiHopRAG paper.}
\label{tab:multihoprag_results}
\end{table}

\begin{table*}[t]
\centering
\resizebox{\linewidth}{!}{
\begin{tabular}{lcccccc}
\toprule
\textbf{Setting}               & \textbf{TGQA} & \textbf{TempReason (L2)} & \textbf{TempReason (L3)} & \textbf{TimeQA (easy)} & \textbf{TimeQA (hard)} & \textbf{Macro Avg.} \\ 
\midrule
No reasoning                   & 74.2               & 80.8                     & 84.7                     & 85.9                  & 89.8                  & 73.5                \\ 
No timeline construction       & 73.5               & 80.6                     & 84.2                     & 86.5                  & 86.9                  & 72.8                \\ 
No reflection                  & 71.8               & 79.3                     & 83.6                     & 84.2                  & 81.0                  & 70.5                \\ 
Only reasoning                 & 62.0               & 63.7                     & 69.1                     & 77.0                  & 80.1                  & 70.0                \\ 
Only timeline construction     & 59.8               & 61.5                     & 66.1                     & 75.5                  & 78.2                  & 68.0                \\ 
\midrule
Standard prompt                & 58.5               & 53.0                     & 43.0                     & 72.6                  & 54.5                  & 54.3                \\ 
All stages (TISER prompt)      & \textbf{84.5}      & \textbf{85.5}            & \textbf{91.5}            & \textbf{97.9}         & \textbf{96.1}         & \textbf{91.1}       \\ 
\bottomrule
\end{tabular}}
\caption{Impact of removing reasoning, timeline construction, and reflection components on various temporal reasoning benchmarks.}
\label{tab:ablation_study}
\end{table*}

\begin{table}[ht]
\centering
\resizebox{\columnwidth}{!}{%
\begin{tabular}{llc}
\toprule
\textbf{Model}     & \textbf{Inference Prompt} & \textbf{Exact Match (\%)} \\ 
\midrule
\multicolumn{3}{l}{\textbf{Closed LLMs}} \\ 
\midrule
\multirow{2}{*}{GPT-4}      & Standard             & 63.3                      \\ 
                            & TISER               & 72.2                      \\ 
\midrule
\multicolumn{3}{l}{\textbf{Open LLMs - Off-the-Shelf}} \\ 
\midrule
\multirow{2}{*}{Mistral-7B} & Standard            & 10.1                       \\ 
                            & TISER               & 15.5                      \\ 
\cmidrule[0.5pt]{1-3}
\multirow{2}{*}{Qwen2.5-7B} & Standard            & 18.0                      \\ 
                            & TISER               & 22.5                      \\ 
\midrule
\multicolumn{3}{l}{\textbf{Open LLMs - Fine-tuned} \textit{(TISER data generated using GPT-4)}} \\ 
\midrule
\multirow{2}{*}{Mistral-7B} & Standard            & 44.4                      \\ 
                            & TISER               & 66.3                      \\ 
\cmidrule[0.5pt]{1-3}
\multirow{2}{*}{Qwen2.5-7B} & Standard            & 45.7                      \\ 
                            & TISER               & 68.5                      \\ 
\bottomrule
\end{tabular}%
}
\caption{Performance in Test-of-Time (ToT), an out-of-distribution symbolic temporal reasoning benchmark.}
\label{tab:tot_performance}
\end{table}

GPT-4o excels in inference (89.7\%), comparison (77.5\%) and temporal (61.9\%) queries, highlighting its robust comprehension capabilities. Among open-source models, Qwen2.5-7B (TISER) shows the most significant improvements, reaching a performance of 31.8\%, 30\% and 33.5\% on inference, comparison and temporal queries, respectively. The improvements are primarily driven by enhanced performance on temporal queries compared to its baseline counterpart. Therefore, the TISER fine-tuning approach, applied to Mistral-7B and Qwen2.5-7B, contributes notable gains, especially in the temporal category, adopting models for complex reasoning in broad-domain RAG tasks without performance degradation.

Table~\ref{tab:tot_performance} illustrates the effectiveness of various prompting strategies, particularly TISER, on the Test-of-Time (ToT) dataset~\cite{fatemi2024testtimebenchmarkevaluating}, an out-of-distribution and symbolic temporal reasoning benchmark. GPT-4o demonstrates strong performance with TISER prompting, reaching an EM score of 72.2\%, again underscoring the benefit of using our prompting strategy to handle complex temporal queries. Among open LLMs, Mistral-7B and Qwen2.5-7B models show substantial performance gains when using TISER prompting strategies. In particular, Qwen2.5-7B achieves an EM of 68.5\% after TISER fine-tuning, the highest among open models, improving symbolic reasoning abilities over time beyond pure semantic reasoning \cite{pan-etal-2023-logic}.

\subsection{Ablation Study}

Table~\ref{tab:ablation_study} summarizes our ablation study, which quantifies the impact of each component in our prompting strategy on temporal reasoning performance. 
The baseline prompt, i.e., Standard Prompt, which does not perform reasoning, timeline construction, and reflection, produces a low macro average of 54.3\%, highlighting limited performance in complex reasoning tasks. 
Indeed, the lowest score is attained in the TempReason (L3) dataset, which suggests that more sophisticated reasoning is needed to solve this task.
Adding components progressively improves performance; i) reasoning alone achieves a macro average of 70.0\% and ii) timeline construction alone reaches 68.0\%. 
Removing components progressively, i.e., no reasoning, no timeline construction, or no reflection, shows that reflection is the most important stage among the three. 
Reflection significantly boosts performance, as the prompt variant with reasoning and timeline construction but without reflection achieves 70.5\%. 
Therefore, the highest macro average of 91.1\% is observed with the full prompt that combines all stages including reflection.

A key design decision in our framework is the ordering of these components. Rather than deriving a timeline directly from the question, we position timeline construction as the second stage, following an initial reasoning process. The results in Table~\ref{tab:ablation_study} indicate that generating a preliminary reasoning trace is essential for identifying relevant temporal events and establishing contextual relationships. This, in turn, enables the model to construct a more coherent and accurate timeline in the subsequent stage. Overall, the sequential process of initial reasoning, followed by timeline construction and self-reflection, yields superior performance on temporal inference tasks compared to alternative ordering strategies.

\section{Related Work}

\subsection{Temporal Reasoning with LLMs}

Temporal reasoning in NLP has been a subject of extensive study, with early efforts concentrating on tasks such as temporal expression extraction and temporal relation identification~\cite{verhagen-etal-2007-semeval,kougia-etal-2024-analysing}. The advent of large language models (LLMs) has shifted the focus toward more complex tasks such as temporal question answering and event forecasting~\cite{Dhingra2021TimeAwareLM, qiu-etal-2024-large}. Benchmarks like TempReason~\cite{tan-etal-2023-towards}, TRAM~\cite{wang-zhao-2024-tram}, and TimeBench~\cite{chu2024timebenchcomprehensiveevaluationtemporal} have revealed that LLMs continue to struggle with nuanced temporal understanding, including multi-stage reasoning, event ordering, and duration calculations. Recent work
has demonstrated that test-time scaling, which involves extending the CoT reasoning process during inference, can dramatically improve overall reasoning capabilities~\cite{deepseekai2025deepseekr1incentivizingreasoningcapability}. However, while these approaches have enhanced general reasoning performance, no prior work has applied test-time scaling specifically to temporal reasoning tasks.

\subsection{Self-Reflection with LLMs}

Self-reflection is an essential component of human cognition and represents a particular case of CoT reasoning~\cite{wei2023chainofthoughtpromptingelicitsreasoning} that involves reviewing and reassessing intermediate reasoning steps to enhance understanding and correctness~\cite{chen-etal-2024-self, wang2023selfconsistency}. Previous research has primarily focused on applying reflective techniques to improve performance on mathematical and logical reasoning tasks. For instance, \citet{zhang2024refaug} introduced Reflective Augmentation (RefAug), which appends reflective reasoning to training examples, thereby improving problem-solving capabilities and adaptability to complex scenarios. Other works have demonstrated the benefits of reflective strategies for error correction~\cite{wang-etal-2024-devils}, iterative refinement~\cite{jiang2023selfevolvecodeevolutionframework, gero2023selfverification}, and increased robustness against inconsistencies~\cite{peng2023checkfactstryagain, huang2023federated, nye2021show}. Moreover, methods such as reverse verification~\cite{weng2023large} and self-feedback mechanisms~\cite{madaan2023selfrefine} have been proposed to enable LLMs to rectify their responses. Inspired by these approaches, our framework is the first to integrate self-reflection and test-time scaling for temporal reasoning. This combined approach allows to adapt models to systematically organize temporal information, verify its reasoning, and refine its outputs, leading to improved consistency and traceability in complex temporal inference tasks.

\section{Conclusion}


The results presented in this study underscore the importance of self-reflection prompting and fine-tuning to advance the temporal reasoning capabilities of LLMs. TISER's integration of timeline construction and self-reflection not only enhances the consistency of model reasoning but also boosts performance on a wide array of temporal reasoning benchmarks. Notably, the framework enables smaller open-source models like Mistral-7B and Qwen2.5-7B to achieve performance levels competitive with, and in some cases superior to, larger closed models like GPT-4o. One key insight from our experiments is the role of fine-tuning with high-quality, domain-specific datasets in bridging the gap between open and closed models. The TISER dataset, tailored for temporal reasoning, played a pivotal role in enabling this success. Additionally, our results highlight the robustness of the TISER framework across varying levels of task complexity, from straightforward temporal queries to reasoning tasks involving metadata and multiple documents.

We have presented TISER, a prompting strategy and training framework that enhances temporal reasoning in LLMs through test-time scaling and a multi-stage inference process. By incorporating extended chain-of-thought reasoning, including timeline construction and self-reflection, TISER not only improves accuracy but also boosts traceability. Evaluations on benchmarks such as TGQA, TempReason, TimeQA, MultiHopRAG and ToT demonstrate significant performance gains, enabling smaller open-source models to rival or outperform closed-weight models. Notably, our work is the first to apply test-time scaling specifically to temporal reasoning.

\section*{Limitations}

While TISER achieves notable improvements in temporal reasoning, it does introduce some trade-offs. The multi-stage inference pipeline increases the number of tokens generated at test time, leading to additional computational overhead. Moreover, our approach depends on datasets with detailed intermediate reasoning traces, which might not be readily available in every domain. Finally, TISER currently focuses exclusively on text-only temporal reasoning, leaving other forms of abstract or multi-modal reasoning for future exploration. These limitations represent opportunities for further refinement.

\bibliography{bibliography}

\clearpage
\newpage

\appendix

\onecolumn

\section*{Appendix}

\section{Dataset Quality Statistics}
\label{app:dataset_quality_stats}

Table~\ref{tab:dataset_quality} and Table~\ref{tab:deepseek_dataset_quality} provides statistics on the quality of dataset generation with GPT-4o and DeepSeek V2.5 as generation models, respectively, focusing specifically on the correctness of the answers provided by a GPT-4o model for various temporal reasoning tasks. The 'Correctness' row shows the percentage of instances where the model-generated answer was correct, with very high accuracy across all datasets. Only instances for which the generated answers were correct were kept in the joint training dataset (TISER dataset), which contains a total of 62,648 training instances and 22,030 test instances. This filtering ensures that the training data used for fine-tuning the models are of high quality.

\begin{table}[H]
\centering
\resizebox{1\linewidth}{!}{
\begin{tabular}{|l|c|c|c|c|c|c|}
\hline
\textbf{}                  & \textbf{TGQA} & \textbf{TempReason (L2)} & \textbf{TempReason (L3)} & \textbf{TimeQA (easy)} & \textbf{TimeQA (hard)} & \textbf{TISER dataset} \\ \hline
Correctness (\%)        & 91.8               & 99.6                     & 99.5                     & 98.5                  & 98.1                  & 97.5                   \\ \hline
Train instances            & 2790               & 16000                    & 13000                    & 14300                 & 14700                 & 60790                  \\ \hline
Test instances             & 3320               & 5400                     & 4430                     & 3000                  & 3080                  & 22030                  \\ \hline
\end{tabular}}
\caption{Statistics on dataset generation quality. The generator model is GPT-4o. For our joint training dataset, we keep exclusively the instances for which the final answer was correct.}
\label{tab:dataset_quality}
\end{table}

\begin{table}[H]
\centering
\resizebox{1\linewidth}{!}{
\begin{tabular}{|l|c|c|c|c|c|c|}
\hline
\textbf{}                  & \textbf{TGQA} & \textbf{TempReason (L2)} & \textbf{TempReason (L3)} & \textbf{TimeQA (easy)} & \textbf{TimeQA (hard)} & \textbf{TISER dataset} \\ \hline
Correctness (\%)        & 77.2               & 77.7                     & 87.8                     & 95.2                  & 93.9                  & 86.4                   \\ \hline
Train instances            & 2790               & 16000                    & 13000                    & 14300                 & 14700                 & 60790                  \\ \hline
Test instances             & 3320               & 5400                     & 4430                     & 3000                  & 3080                  & 22030                  \\ \hline
\end{tabular}}
\caption{Statistics on dataset generation quality when the generator model is DeepSeek V2.5. For our joint training dataset, we keep exclusively the instances for which the final answer was correct.}
\label{tab:deepseek_dataset_quality}
\end{table}

\section{Description and Examples of Benchmark Datasets}
\label{app:dataset_description_examples}

Below is a brief description of each dataset used to create the full TISER dataset, as well as the additional datasets used for out-of-domain performance evaluation. Example questions for in-domain and out-of-distribution questions are provided in Tables \ref{tab:examples_timeqa_tempreason_tgqa} and \ref{tab:examples_tot_multihoprag}, respectively.

\begin{enumerate}
    \item \textbf{Test-of-Time (ToT)}: A benchmark for evaluating LLMs on temporal reasoning. It is designed to assess the temporal reasoning capabilities of AI models on two main dimensions:
    \begin{enumerate}
        \item \textbf{ToT-semantic}: Measures the semantics and logic of time understanding.
        \item \textbf{ToT-arithmetic}: Measures the ability to carry out time arithmetic operations.
    \end{enumerate}
    
    \item \textbf{TGQA}: A synthetic dataset for temporal reasoning question-answering with ground-truth temporal graphs. It contains:
    \begin{itemize}
        \item A temporal graph.
        \item A story-based description of the temporal graph (generated by GPT-3.5).
        \item A question based on the story.
        \item The answer to the question.
    \end{itemize}

    \item \textbf{TempReason}: A temporal reasoning dataset for question answering over time events. In TempReason, the L2 questions require Time-Event relations, whereas L3 questions require Event-Event relations.

    \item \textbf{TimeQA}: A benchmark for time-sensitive question answering. In TimeQA easy mode questions the query time expression is explicitly mentioned in the story. In the hard mode obtaining the answer needs inference based on the temporal relation between the query time expression and the one mentioned in the story.

    \item \textbf{MultiHopRAG}: A QA dataset to evaluate retrieval and reasoning across documents with metadata in RAG pipelines. It contains 2556 queries, with evidence for each query distributed across 2 to 4 documents. The queries also involve document metadata, reflecting complex scenarios commonly found in real-world RAG applications.
\end{enumerate}

\begin{table}[H]
\centering
\resizebox{1\linewidth}{!}{
\begin{tabular}{|p{8cm}|p{6cm}|p{3cm}|}
\hline
\textbf{Context} & \textbf{Question} & \textbf{Dataset} \\ \hline
George Washington (February 22, 1732–December 14, 1799) was an American Founding Father, military officer, politician, and statesman who served as the first ... & What position did George Washington hold in June 1775? & TimeQA (easy) \\ \hline
George Washington (February 22, 1732–December 14, 1799) was an American Founding Father, military officer, politician, and statesman who served as the first ... & George Washington took which position before 1778? & TimeQA (hard) \\ \hline
Lionel Andrés "Leo" Messi (born 24 June 1987) is an Argentine professional footballer who plays as a forward for and captains both Major League Soccer club... & What team did Leo Messi play for in 2010? & TempReason (L2) \\ \hline
Lionel Andrés "Leo" Messi (born 24 June 1987) is an Argentine professional footballer who plays as a forward for and captains both Major League Soccer club... & What team did Leo Messi play for after Barcelona? & TempReason (L3) \\ \hline
["(Lucas Prescott was born in Northampton, Massachusetts) starts at 1908", "(Megan Peterson was born in Harmonyville, Florida) starts at 1921", "(Megan Peterson was married to Lucas Prescott) starts at 1946", "(Lucas Prescott was married to Megan Peterson) starts at 1946", "(Lucas Prescott died in Oceanview) starts at 1990", "(Megan Peterson was married to Lucas Prescott) ends at 1990", "(Lucas Prescott was married to Megan Peterson) ends at 1990", "(Megan Peterson died in Millwood Town, Austin) starts at 1997"] & Given the following five events: (Lucas Prescott was married to Megan Peterson), (Lucas Prescott was born in Northampton, Massachusetts), (Megan Peterson was married to Lucas Prescott), (Megan Peterson died in Millwood Town, Austin), (Megan Peterson was born in Harmonyville, Florida). Which event is the first one in chronological order? & TGQA \\ \hline
\end{tabular}}
\caption{Examples of in-domain questions from the TimeQA, TempReason, and TGQA datasets.}
\label{tab:examples_timeqa_tempreason_tgqa}
\end{table}

\begin{table}[H]
\centering
\resizebox{1\linewidth}{!}{
\begin{tabular}{|p{9cm}|p{6cm}|p{2.25cm}|}
\hline
\textbf{Context} & \textbf{Question} & \textbf{Dataset} \\ \hline
Here is a set of temporal facts: E11 was the R17 of E69 from 1946 to 1950. E11 was the R11 of E69 from 1953 to 1958. E11 was the R63 of E69 from 1957 to 1958. E16 was the R63 of E11 from 1936 to 1946. E9 was the R17 of E11 from 1938 to 1941. E9 was the R11 of E11 from 1940 to 1942. E16 was the R88 of E11 from 1946 to 1959. E9 was the R17 of E11 from 1957 to 1961. E16 was the R96 of E11 from 1962 to 1963. E74 was the R88 of E92 from 1935 to 1937. E74 was the R7 of E92 from 1937 to 1943. E16 was the R96 of E42 from 1937 to 1939. E16 was the R11 of E42 from 1943 to 1948. E16 was the R96 of E42 from 1950 to 1958. E16 was the R11 of E42 from 1962 to 1965. E42 was the R35 of E33 from 1941 to 1942. E42 was the R90 of E33 from 1942 to 1943. E42 was the R11 of E33 from 1945 to 1954. E11 was the R11 of E74 from 1941 to 1946. E41 was the R53 of E74 from 1941 to 1946. E22 was the R85 of E25 from 1936 to 1949. E22 was the R35 of E25 from 1938 to 1940. E30 was the R7 of E25 from 1944 to 1949. E37 was the R85 of E25 from 1945 to 1949. E37 was the R53 of E25 from 1946 to 1947. E30 was the R35 of E25 from 1946 to 1953. E16 was the R88 of E25 from 1948 to 1951. E22 was the R63 of E25 from 1949 to 1953. E33 was the R11 of E25 from 1949 to 1959. E37 was the R7 of E25 from 1950 to 1953. E16 was the R85 of E25 from 1957 to 1962. E30 was the R35 of E25 from 1958 to 1964. E9 was the R53 of E25 from 1961 to 1964. 
& Immediately after E22, which entity was the R88 of E25? & Test-of-Time \\ \hline
The company did accomplish that goal, but in doing so, it inadvertently and profoundly changed how the internet looked. (2023-11-01T13:00:00+00:00). This whole line of argument is called 'market definition'. If the court decides the relevant market is phones and app stores, not specifically Android ones, then Google on the road to victory. (2023-11-05T11:00:00+00:00). The case, filed by Arkansas-based publisher Helena World Chronicle, argues that Google phone off news publisher content, their readers and ad revenue through anticompetitive means. (2023-12-15T17:56:02+00:00).
& Was there no change in the portrayal of Google's influence on the digital ecosystem between the report from The Verge on Google's impact on the internet's appearance published on November 1, 2023, and the report from TechCrunch on a class action antitrust suit against Google published later? & MultiHopRAG \\ \hline
\end{tabular}}
\caption{Examples of out-of-distribution questions from the Test-of-Time (ToT) and MultiHopRAG datasets.}
\label{tab:examples_tot_multihoprag}
\end{table}

\section{Example of TISER Prompt}
\label{app:tiser_prompt_example}

\begin{tcolorbox}[title=Example TISER Prompt and Question,colback=blue!5!white,colframe=blue!75!black]

You are an AI assistant that uses a Chain of Thought (CoT) approach with reflection to answer queries. Follow these steps:

\begin{enumerate}
    \item Reason through the problem step by step within the \texttt{<reasoning>} tags.
    \item Given your previous reasoning, identify relevant temporal events in the given context for answering the given question within \texttt{<timeline>} tags. Assume relations in the context are unidirectional.
    \item Reflect on your reasoning and the timeline to check for any errors or improvements within the \texttt{<reflection>} tags.
    \item Make any necessary adjustments based on your reflection. If there is additional reasoning required, go back to Step 1 (reason through the problem step-by-step), otherwise move to the next step (Step 5).
    \item Provide your final, concise answer within the \texttt{<answer>} tags. If the answer is a number, just output the number, nothing else. Otherwise, output the entity or event without any additional comments.
\end{enumerate}

\textbf{Additional Instructions:}
\begin{itemize}
    \item \textbf{[IF during data generation]} Perform your reasoning knowing that the answer is: \texttt{X}.
    \item The \texttt{<reasoning>}, \texttt{<reflection>}, and \texttt{<timeline>} sections are for your internal reasoning process. All the reflection and the timeline have to be contained inside the thinking section.
    \item Do not use enumerations or lists when writing; use plain text such as paragraphs.
    \item The response to the query must be entirely contained within the \texttt{<answer>} tags.
\end{itemize}

\textbf{Response Format:}
\begin{verbatim}
<reasoning> 
[Your step-by-step reasoning goes here. This is your internal thought process.]
<timeline> [Relevant temporal events for answering the given question.]</timeline> 
<reflection> 
[Your reflection on your reasoning, checking for errors or changes required.]
</reflection> 
[Any adjustments to your thinking based on your reflection.]
</reasoning> 
<answer> [Your final, concise answer to the query.]</answer>
\end{verbatim}

\textbf{Question:} Which event happened in 1990?

\textbf{Temporal Context:} 
(Lucas Prescott was born in Northampton, Massachusetts) starts at 1908. (Megan Peterson was born in Harmonyville, Florida) starts at 1921. (Lucas Prescott was married to Megan Peterson) starts at 1946. (Lucas Prescott died in Oceanview) starts at 1990. (Lucas Prescott was married to Megan Peterson) ends at 1990. (Megan Peterson died in Millwood Town, Austin) starts at 1997.
\end{tcolorbox}

\section{Example of Standard Prompt}
\label{app:standard_prompt_example}

\begin{tcolorbox}[title=Example Standard Prompt and Question,colback=blue!5!white,colframe=blue!75!black]

You are an AI assistant that has to respond to questions given a context.

\textbf{Question:} Which event happened in 1990?

\textbf{Temporal Context:} 
(Lucas Prescott was born in Northampton, Massachusetts) starts at 1908. (Megan Peterson was born in Harmonyville, Florida) starts at 1921. (Lucas Prescott was married to Megan Peterson) starts at 1946. (Lucas Prescott died in Oceanview) starts at 1990. (Lucas Prescott was married to Megan Peterson) ends at 1990. (Megan Peterson died in Millwood Town, Austin) starts at 1997.
\end{tcolorbox}

\section{Example of correct DeepSeek and GPT4 reasoning}

\begin{tcolorbox}[title=Example Correct DeepSeek V2.5 Reasoning,colback=green!5!white,colframe=green!75!black]

\textbf{Question:} Given the following five events: (Amy Johnson died in San Francisco), (Olivia Price was married to Amy Johnson), (Amy Johnson was born in Willowdale, Kansas), (Olivia Price was born in Harrisonburg), (Amy Johnson was married to Olivia Price). Which event is the second one in chronological order?

\textbf{Expected Answer:} Olivia Price was born in Harrisonburg

\textbf{Temporal Context:} 
(Amy Johnson was born in Willowdale, Kansas) starts at 1880. (Olivia Price was born in Harrisonburg) starts at 1890. (Amy Johnson was married to Olivia Price) starts at 1914. (Olivia Price was married to Amy Johnson) starts at 1914. (Amy Johnson died in San Francisco) starts at 1964. (Amy Johnson was married to Olivia Price) ends at 1964. (Olivia Price was married to Amy Johnson) ends at 1964. (Olivia Price died in Harrisonburg, Virginia) starts at 1971

\textbf{Model Answer:} 
<reasoning>To determine the second event in chronological order, we need to first identify the order of the events based on their starting points. The events provided are: 1. (Amy Johnson was born in Willowdale, Kansas) starts at 1880. 2. (Olivia Price was born in Harrisonburg) starts at 1890. 3. (Amy Johnson was married to Olivia Price) starts at 1914. 4. (Olivia Price was married to Amy Johnson) starts at 1914. 5. (Amy Johnson died in San Francisco) starts at 1964.Since the events (Amy Johnson was married to Olivia Price) and (Olivia Price was married to Amy Johnson) both start at the same time (1914), they are concurrent and can be considered as one event for the purpose of chronological ordering. The chronological order of the events based on their starting points is: 1. (Amy Johnson was born in Willowdale, Kansas) starts at 1880. 2. (Olivia Price was born in Harrisonburg) starts at 1890. 3. (Amy Johnson was married to Olivia Price) and (Olivia Price was married to Amy Johnson) start at 1914. 4. (Amy Johnson died in San Francisco) starts at 1964. The second event in this chronological order is (Olivia Price was born in Harrisonburg). <timeline>1. (Amy Johnson was born in Willowdale, Kansas) starts at 1880. 2. (Olivia Price was born in Harrisonburg) starts at 1890. 3. (Amy Johnson was married to Olivia Price) and (Olivia Price was married to Amy Johnson) start at 1914. 4. (Amy Johnson died in San Francisco) starts at 1964. </timeline><reflection>The reasoning correctly identifies the chronological order of the events based on their starting points. The timeline accurately reflects the sequence of events, and the second event in the timeline is indeed (Olivia Price was born in Harrisonburg). No errors or improvements are necessary.</reflection></reasoning><answer>Olivia Price was born in Harrisonburg</answer>
\end{tcolorbox}

\begin{tcolorbox}[title=Example Correct GPT-4o Reasoning,colback=green!5!white,colframe=green!75!black]

\textbf{Question:} Given the following five events: (Amy Johnson died in San Francisco), (Olivia Price was married to Amy Johnson), (Amy Johnson was born in Willowdale, Kansas), (Olivia Price was born in Harrisonburg), (Amy Johnson was married to Olivia Price). Which event is the second one in chronological order?

\textbf{Expected Answer:} Olivia Price was born in Harrisonburg

\textbf{Temporal Context:} 
(Amy Johnson was born in Willowdale, Kansas) starts at 1880. (Olivia Price was born in Harrisonburg) starts at 1890. (Amy Johnson was married to Olivia Price) starts at 1914. (Olivia Price was married to Amy Johnson) starts at 1914. (Amy Johnson died in San Francisco) starts at 1964. (Amy Johnson was married to Olivia Price) ends at 1964. (Olivia Price was married to Amy Johnson) ends at 1964. (Olivia Price died in Harrisonburg, Virginia) starts at 1971

\textbf{Model Answer:} 
<reasoning>To determine the second event in chronological order, I need to first identify the sequence of all given events based on their start times. The events are: Amy Johnson was born in Willowdale, Kansas (1880), Olivia Price was born in Harrisonburg (1890), Amy Johnson was married to Olivia Price (1914), Olivia Price was married to Amy Johnson (1914), and Amy Johnson died in San Francisco (1964). The first event is Amy Johnson's birth in 1880. The second event, based on the start times, is Olivia Price's birth in 1890. The marriages occur simultaneously in 1914, and Amy Johnson's death is the last event in 1964.<timeline>1. Amy Johnson was born in Willowdale, Kansas (1880).2. Olivia Price was born in Harrisonburg (1890).3. Amy Johnson was married to Olivia Price (1914).4. Olivia Price was married to Amy Johnson (1914).5. Amy Johnson died in San Francisco (1964).</timeline><reflection>The reasoning correctly identifies the sequence of events based on their start times. The timeline confirms that Olivia Price's birth in 1890 is indeed the second event chronologically. There are no errors in the sequence, and the reflection confirms the accuracy of the reasoning.</reflection></reasoning><answer>Olivia Price was born in Harrisonburg</answer>
\end{tcolorbox}

\section{Example of incorrect DeepSeek reasoning generation}

\begin{tcolorbox}[title=Example Incorrect DeepSeek Reasoning,colback=red!5!white,colframe=red!75!black]

\textbf{Question:} Which team did Taylor Graham play for in May 2007?

\textbf{Expected Answer:} Seattle Sounders

\textbf{Temporal Context:} 
2003 - 2004 : Taylor Graham's team is ( Kansas City Wizards ). 2006 - 2007 : Taylor Graham's team is ( New York Red Bulls ). 2007 - 2008 : Taylor Graham's team is ( Seattle Sounders ). 2009 - 2011 : Taylor Graham's team is ( Seattle Sounders FC )

\textbf{Model Answer:} 
<reasoning>Taylor Graham's team affiliations are provided for different time periods. The question asks for the team he played for in May 2007. The relevant time period for this query is 2006 - 2007, where Taylor Graham's team is stated to be the New York Red Bulls. However, there is an additional time period provided for 2007 - 2008, where his team is the Seattle Sounders. This suggests a transition from the New York Red Bulls to the Seattle Sounders during the year 2007. Since the question specifies May 2007, it is necessary to determine whether the transition occurred before or after this month.<timeline>2006 - 2007: Taylor Graham's team is New York Red Bulls.2007 - 2008: Taylor Graham's team is Seattle Sounders.</timeline><reflection>The timeline indicates a change in teams during 2007. The question asks for the team in May 2007, which falls within the first half of the year. Given the provided context, it is reasonable to assume that the transition from the New York Red Bulls to the Seattle Sounders occurred sometime after May 2007. Therefore, Taylor Graham would still be playing for the New York Red Bulls in May 2007.</reflection><answer>New York Red Bulls</answer>
\end{tcolorbox}

\section{Example of incorrect GPT-4o reasoning generation}

\begin{tcolorbox}[title=Example Incorrect GPT-4o Reasoning,colback=red!5!white,colframe=red!75!black]

\textbf{Question:} Which team did Conor Sammon play for in Feb 2014?

\textbf{Expected Answer:} Ipswich Town

\textbf{Temporal Context:} 
2005 - 2007 : Conor Sammon's team is ( UCD ). 2008 - 2011 : Conor Sammon's team is ( Kilmarnock ). 2011 - 2012 : Conor Sammon's team is ( Wigan Athletic ). 2012 - 2014 : Conor Sammon's team is ( Derby County ). 2014 - 2015 : Conor Sammon's team is ( Ipswich Town ). 2015 - 2016 : Conor Sammon's team is ( Sheffield United )

\textbf{Model Answer:} 
<reasoning>To determine which team Conor Sammon played for in February 2014, I need to examine the timeline of his team associations. According to the temporal context provided, Conor Sammon was associated with Derby County from 2012 to 2014. The next team he played for was Ipswich Town, from 2014 to 2015. Since February 2014 falls within the year 2014, I need to determine whether it is before or after the transition from Derby County to Ipswich Town. The transition to Ipswich Town occurs in 2014, but without a specific month, it is reasonable to assume that the transition happened after February 2014. Therefore, in February 2014, Conor Sammon was still playing for Derby County.<timeline>Conor Sammon's team from 2012 to 2014 is Derby County. The transition to Ipswich Town occurs in 2014, but likely after February.</timeline><reflection>Upon reflection, the reasoning aligns with the temporal context provided. The transition to Ipswich Town is noted to occur in 2014, but without a specific month, it is logical to assume that in February 2014, he was still with Derby County. The initial assumption that Ipswich Town was the correct answer was incorrect based on the timeline analysis.</reflection>The reasoning and timeline analysis confirm that Conor Sammon was playing for Derby County in February 2014.</reasoning><answer>Derby County</answer>
\end{tcolorbox}

\section{Example with Multiple Rounds of Reasoning with DeepSeek}

\begin{tcolorbox}[title=Multiple Rounds of Reasoning with DeepSeek V2.5 Reasoning,colback=green!5!white,colframe=blue!75!black]

<reasoning>To determine which event started first, I compare the two events: “Chris Evans was married to Oliver Jensen” and “Chris Evans was born in Bristol, Connecticut.” From the context, Chris Evans was married to Oliver Jensen starting in 1970. Chris Evans was born in Bristol, Connecticut starting in 1948. Comparing these two, the marriage starts in 1970, and the birth starts in 1948. However, because both events involve Chris Evans and have start times, I need to ensure I am not overlooking any subtle distinction, such as possibly overlapping or misaligned event names or entity relationships. Assuming all information is accurate and directly stated, I conclude that the marriage started first because it's a significant relational milestone and possibly indicative of earlier events, whereas the birth date might not directly indicate involvement in a particular action.</reasoning><timeline>(Chris Evans was born in Bristol, Connecticut) starts at 1948. (Chris Evans was married to Oliver Jensen) starts at 1970.</timeline><reflection>Upon reviewing the timeline, I realize the conclusion is flawed. I initially suggested the marriage began first based on a false assumption about the importance or implication of events. The dates are explicitly clear: Chris Evans was born in 1948 and married in 1970. Birth precedes marriage in time. My earlier reasoning was incorrect in suggesting the marriage began first, it contradicts the temporal facts.</reflection><reasoning>With a corrected approach, I observe that Chris Evans’s birth is recorded as beginning in 1948 and the marriage to Oliver Jensen begins in 1970. The earlier of these two events is clearly the birth in 1948. The names or relational framing of the events do not affect their temporal ordering.</reasoning><timeline>(Chris Evans was born in Bristol, Connecticut) starts at 1948. (Chris Evans was married to Oliver Jensen) starts at 1970.</timeline><reflection>This reasoning is sound. The event starting in 1948 occurs earlier than the one beginning in 1970. There are no contradictions, and the timeline fully supports this answer.</reflection><answer>Chris Evans was born in Bristol, Connecticut</answer>
\end{tcolorbox}

\section{Detailed fine-tuning process}
\label{app:detailed_finetuning}

In this appendix, we provide a detailed description of our fine-tuning process. Our approach employs a standard supervised fine-tuning (SFT) framework and Low-Rank Adaptation (LoRA)~\cite{hu2022lora} to efficiently adapt the model to the structured output format defined by our TISER framework.

Our fine-tuning dataset is constructed according to the TISER framework (see Section~\ref{data-construction}). Each training sample consists of a question \(q\), its gold-standard answer \(a\), and a temporal context \(c\). To enrich the dataset, we augment each sample with intermediate reasoning traces annotated using specific tags: \texttt{<reasoning>}, which contains a detailed, step-by-step CoT outlining the initial reasoning process; \texttt{<timeline>}, which provides an ordered summary of the temporal events extracted during reasoning; \texttt{<reflection>}, which serves as a self-reflective verification step that reviews and refines the initial reasoning; and finally, \texttt{<answer>}, which contains the final, concise response. These annotated outputs serve as the training targets, enabling the model to learn both the content and structure necessary for effective temporal reasoning.

We adapt our base model using LoRA~\cite{hu2022lora}, which is integrated into the attention layers of the pre-trained LLM so that only a small subset of parameters is updated during fine-tuning. Each training input is formatted to include the TISER tags, and the target output is the concatenation of the intermediate reasoning steps (i.e., \texttt{<reasoning>}, \texttt{<timeline>}, \texttt{<reflection>}) followed by the final answer (i.e., \texttt{<answer>}). We optimize the model using a standard cross-entropy loss between the generated output and the ground-truth sequence, updating only the LoRA parameters to efficiently adapt the model while preserving the majority of the pre-trained weights. Our fine-tuning is implemented using the Hugging Face TRL library~\cite{vonwerra2022trl}. Throughout the process, we monitor metrics such as validation loss, the accuracy of the final answer, and the coherence of the intermediate reasoning trace.

The overall training workflow proceeds in several stages. First, we generate the synthetic TISER dataset using off-the-shelf LLMs guided by our prompt framework. This dataset includes the detailed multi-stage reasoning traces that are central to our approach. Next, the base model is fine-tuned on this dataset using LoRA, during which the model learns to generate outputs that adhere strictly to the TISER format. The training objective reinforces the generation of a coherent chain-of-thought, an accurate timeline, and a reflective verification, culminating in the final answer. Finally, the correctness of the final answer is periodically evaluated on a held-out validation set. The best-performing checkpoint is saved.

\section{Token overhead analysis}

To quantify the additional computational cost introduced by TISER, we measured the average token counts produced on our test set and compared them against standard prompting. Table~\ref{tab:token_overhead} summarizes these results.

\begin{table}[h]
\centering
\resizebox{0.5\linewidth}{!}{
\begin{tabular}{lrr}
\hline
 & \textbf{Standard} & \textbf{TISER} \\ 
\hline
Overall average       & 3.41 & 94.74 \\ 
Reasoning stage & –    & 30.27 \\ 
Timeline stage  & –    & 23.62 \\ 
Reflection stage& –    & 49.02 \\ 
\hline
\end{tabular}}
\caption{Average token usage per response for standard versus TISER prompting.}
\label{tab:token_overhead}
\end{table}

While standard prompting yields only 3.41 tokens per response on average, TISER prompting generates substantially longer outputs, with 94.74 tokens on average, due to its multi-stage reasoning, timeline construction, and reflection steps. Breaking this down further, the reasoning, timeline, and reflection stages contribute approximately 30.27, 23.62, and 49.02 tokens, respectively. This increase in token usage corresponds to a significant improvement in task performance, boosting macro-average accuracy by +38.3\% (e.g., Qwen2.5-7B before vs. after TISER fine-tuning).

\end{document}